\documentclass[preprint,12pt]{elsarticle}

\usepackage{amssymb}

 \usepackage{amsthm}
\usepackage{xcolor}
\usepackage{enumitem}
\usepackage{caption}
\usepackage{subcaption}
\usepackage{longtable}
\usepackage{scrextend}
\usepackage{multirow}
\usepackage{array}
\usepackage{listings}
\usepackage[T1]{fontenc}
\usepackage{tgbonum}
\usepackage{makecell}
\usepackage{textcomp}
\usepackage{tablefootnote}
\usepackage{url}
\usepackage{amsmath}
\usepackage{float}
 
\colorlet{punct}{red!60!black}
\definecolor{background}{HTML}{EEEEEE}
\definecolor{delim}{RGB}{20,105,176}
\colorlet{numb}{magenta!60!black}

\lstdefinelanguage{json}{
    basicstyle=\scriptsize\ttfamily,
    numbers=left,
    numberstyle=\scriptsize,
    stepnumber=1,
    numbersep=8pt,
    showstringspaces=false,
    breaklines=true,
    frame=lines,
    backgroundcolor=\color{background},
    literate=
     *{0}{{{\color{numb}0}}}{1}
      {1}{{{\color{numb}1}}}{1}
      {2}{{{\color{numb}2}}}{1}
      {3}{{{\color{numb}3}}}{1}
      {4}{{{\color{numb}4}}}{1}
      {5}{{{\color{numb}5}}}{1}
      {6}{{{\color{numb}6}}}{1}
      {7}{{{\color{numb}7}}}{1}
      {8}{{{\color{numb}8}}}{1}
      {9}{{{\color{numb}9}}}{1}
      {:}{{{\color{punct}{:}}}}{1}
      {,}{{{\color{punct}{,}}}}{1}
      {\{}{{{\color{delim}{\{}}}}{1}
      {\}}{{{\color{delim}{\}}}}}{1}
      {[}{{{\color{delim}{[}}}}{1}
      {]}{{{\color{delim}{]}}}}{1},
}

\journal{}

\begin{document}

\begin{frontmatter}



\title{The FAIRy Tale of Genetic Algorithms}


\author[uos]{Fahad Maqbool\corref{cor1}}
\ead{fahad.maqbool@uos.edu.pk}
\cortext[cor1]{Corresponding Author}
\address[uos]{University of Sargodha, Pakistan}
\author[uos]{Muhammad Saad Razzaq}
\ead{saad.razzaq@uos.edu.pk}
\author[gesis]{Hajira Jabeen}
\ead{hajira.jabeen@gesis.org}
\address[gesis]{GESIS-Leibniz Institute for the Social Sciences, Cologne, Germany}

\begin{abstract}
Genetic Algorithm (GA) is a popular meta-heuristic evolutionary algorithm that uses stochastic operators to find optimal solution and has proved its effectiveness in solving many complex optimization problems (such as classification, optimization, and scheduling). However, despite its performance, popularity and simplicity, not much attention has been paid towards reproducibility and reusability of GA. 
In this paper, we have extended Findable, Accessible, Interoperable and Reusable (FAIR) data principles to enable the reproducibility and reusability of algorithms. We have chosen GA as a usecase to the demonstrate the applicability of the proposed principles. Also we have presented an overview of methodological developments and variants of GA that makes it challenging to reproduce or even find the right source. Additionally, to enable FAIR algorithms, we propose a vocabulary (i.e. $evo$) using light weight RDF format, facilitating the reproducibility. Given the stochastic nature of GAs, this work can be extended to numerous Optimization and machine learning algorithms/methods.
\end{abstract}



\begin{keyword}
FAIR \sep Genetic Algorithm \sep Metadata \sep Digital Artifact \sep Reproducibility \sep Reusability \sep Evolutionary Algorithm.



\end{keyword}

\end{frontmatter}


\section{Introduction}  \label{sec:Intro}
{T}{raditional} optimization techniques (random search, univariate method, stochastic gradient decent, quasi-Newton) are good at solving simple optimization problems \cite{chambers2019practical,mitchell1998introduction}. However, they suffer from the performance bottleneck with an increase in the problem complexity. Also, these techniques require a well-defined deterministic path at the start. On the other hand, stochastic optimization techniques like GA perform well with non-smooth and ill-conditioned objective functions. It is capable to find good solutions while avoiding local optima. GA is based on the idea of “Survival of the fittest”. Given a population, it has three main operators i.e. Selection, Mutation, and Crossover. Selection chooses potentially promising solutions to proceed to the next generation, while crossover combines the traits of parent chromosomes to create the offsprings. Mutation changes a certain value of a gene within a chromosome, and this helps in avoiding local optima. GA has a wide application range including scheduling, planning, assignment, and prediction in various industry and business problems \cite{liu2015scalable, trivedi2015hybridizing}.\\
Currently, we are in the middle of the golden jubilee and diamond jubilee of Genetic Algorithms but far away from the standardization of GA. Even after 50+ years, we are unable to decide and agree on the name that corresponds to a particular set of hyperparameters. Genetic Algorithm was the term coined by John Holland in 1965 \cite{chambers2019practical,mitchell1998introduction}. Since then, Genetic Algorithms \cite{chambers2019practical}, Simple Genetic Algorithms \cite{mitchell1998introduction}, Canonical Genetic Algorithms
\cite{chambers2019practical}, and Sequential Genetic Algorithms \cite{mitchell1998introduction}
 are the several names given to GA. One might be confused if these are different GA variants, but these all are the same and refer to John Holland's GA. Similar is the case with GA software (source code of a GA research publication). One may find different code repositories of GA\footnote{\label{1st}\url{https://github.com/bz51/GeneticAlgorithm}} \footnote{\label{2nd}\url{ https://github.com/ezstoltz/genetic-algorithm}}, Simple GA Simple Genetic Algorithm\footnote{\label{3rd} \url{https://github.com/tmsquill/simple-ga}} \footnote{\label{4th}\url{https://github.com/yetanotherchris/SimpleGeneticAlgorithm}} 
\footnote{\label{5th}\url{https://github.com/afiskon/simple-genetic-algorithm}}
\footnote{\label{6th}\url{https://github.com/ajlopez/SimpleGA}}, Canonical Genetic Algorithm\footnote{\label{7th}\url{https://github.com/GMTurbo/canonical-ga}} \footnote{\label{8th}\url{https://github.com/nanoff/Canonical-Genetic-Algorithm}} \footnote{\label{9th}\url{https://github.com/sanamadanii/Canonical-Genetic-Algorithm}} \footnote{\label{10th}\url{https://github.com/sajjadaemmi/Canonical-Genetic-Algorithm}} \footnote{\label{11}\url{https://github.com/yareddada/Canonical-Genetic-Algorithm}}
\footnote{\label{12}\url{https://github.com/UristMcMiner/canonical_genetic_algorithm}}
, and Sequential Genetic Algorithm\footnote{\label{13}\url{https://github.com/regicsf2010/SequentialGA}}
 on GitHub\footnote{\label{14}\url{https://guides.github.com/features/pages/}}.These various implementations of the same approach with different naming conventions may decrease the findability, accessibility and reusability of GA. Also, one may find various implementations of the GA\footnote{\url{https://github.com/ezstoltz/genetic-algorithm}} \footnote{\url{https://github.com/strawberry-magic-pocket/Genetic-Algorithm}} \footnote{\url{https://github.com/memento/GeneticAlgorithm}} \footnote{\url{https://github.com/ShiSanChuan/GeneticAlgorithm}} \footnote{\url{https://github.com/lagodiuk/genetic-algorithm}} with different hyperparameters but with the same naming conventions. This make it difficult in understandability and reusability.\\
  
A motivation behind this work is that a considerable portion of scientific data and research manuscripts remain unnoticed every year due to partial findability, accessibility, reusability, and interoperability by humans or machines \cite{stall2019make,wilkinson2016FAIR,vogt2019FAIR,wilkinson2018design,guizzardi2020ontology,vogt2019anatomy,fantin2020distributed,hasnain2018assessing}. Only one-fifth of the published manuscripts also publish experimental data on some data repositories \cite{stall2019make}. In current research practices, most of the data used in research articles is not findable. Hence, it cannot easily be reused by the research community. Similarly, the act of sharing research software (i.e. code and hyper-parameter settings) is not a common practice due to little to no attribution mechanisms for the research software developers 
\cite{hasselbring2020FAIR}. In most cases, research software's details are very briefly shared in research manuscripts and hence are far from being findable and reproducible. Same naming conventions of different digital artifacts, different naming convention of same digital artifact, ignored practices of publishing dataset and software code with research articles, sharing code in repositories without rich metadata adds a challenge to code findability and hence compromises the algorithm reusability and reproducibility. 
In recent years, efforts have been made in research, academia, development, and industry to make scientific data FAIR (Findable, Accessible, Interoperable, Reusable) for both humans and machines \cite{wilkinson2016FAIR,stall2019make}. Table \ref{tab:FAIR for data} briefly covers FAIR data principles proposed by Wilkinson et al. \cite{wilkinson2016FAIR} in 2016 for making data FAIR. These principles focus on machine action-ability with minimum to nil human intervention.

\begin{table*}[!htb]
\scriptsize
\caption{FAIR data principles\label{tab:FAIR for data}}
\renewcommand{\arraystretch}{1.1}

\centering
\begin{tabular}{|p{0.05\linewidth}|p{0.05\linewidth}|p{0.8\linewidth}|}
\hline
FAIR & Id & Description \\
\hline

\multirow{4}{*}{F} & {1} & metadata are assigned a globally unique and persistent identifier. \\
\cline{2-3}
 & {2} & data are described with rich metadata. \\
\cline{2-3}
 & {3} & metadata clearly and explicitly include the identifier of the data it describes. \\
\cline{2-3}
  & {4} & metadata are registered or indexed in a searchable resource.\\
\hline

\multirow{4}{*}{A} & {1} & metadata are retrievable by their identifier using a standardized Communications protocol.    \\
\cline{2-3}
 & {1.1} & the protocol is open, free, and universally implementable.\\
\cline{2-3}
 & {1.2} & the protocol allows for an authentication and authorization procedure, where necessary.\\
\cline{2-3}
 & {2} & metadata are accessible, even when the data are no longer available.   \\
\hline

\multirow{3}{*}{I} & {1} & metadata use a formal, accessible, shared, and broadly applicable language to facilitate machine readability and data exchange.  \\
\cline{2-3}
 & {2} & metadata use vocabularies that follow FAIR principles.  \\
\cline{2-3}
 & {3} & metadata include qualified references to other (meta)data.  \\
\hline

\multirow{4}{*}{R} & {1} & metadata are richly described with a plurality of accurate and relevant attributes. \\
\cline{2-3}
 & {1.1} & metadata are released with a clear, and accessible data usage license.\\
\cline{2-3}
 & {1.2} & metadata are associated with detailed provenance. \\
\cline{2-3}
 & {1.3} & metadata meet domain-relevant community standards. \\
\hline

\end{tabular}
\end{table*}

FAIR principles revolve around three main components (i.e Digital Artifact, Metadata about the digital artifact and Infrastructure). The FAIR guidelines emphasize automated discovery (Findability) of the digital artifact (mainly data). Once discovered one should have a clear idea of how these artifacts can be accessed including authentication and authorization. Metadata should be well defined to assist reusability. Data is more productive if its accessibility, interoperability, and reusability details are clearly documented in its metadata.\\  
The contribution of this article is\\
\begin{enumerate}[noitemsep]
\item We have extended FAIR principles beyond data, so that these could be applied to methods, algorithms and software artifacts. 
\item We have presented GA as a usecase to demonstrate the applicability of proposed FAIR principles for algorithms.
\item We have proposed specialized metadata for GA to ensure FAIR practice using light weight RDF format.
\item We demonstrate the application of proposed principles for a Python based GA code \footnote{\label{PC}\url{https://doi.org/10.5281/zenodo.7096663}} and published its associated metadata \footnote{\label{mt}\url{https://doi.org/10.5281/zenodo.7095155}} through zenodo.

\end{enumerate}

The rest of the article is comprising of section \ref{sec:Prelim} that highlights the preliminaries of GA, FAIR principles, and pointed the challenges currently being faced by the research community. In section \ref{sec:Related Work} we have explored the relevant literature and summarized the recent development on FAIR and highlighted the challenges of fostering the FAIR culture. Section \ref{sec:FAIR Principles} covers the FAIR common and exclusive principles for algorithms and GA, while section \ref{sec:GA-Metadata} presents the metadata of GA. Mapping of FAIR Algorithms principles on GA is highlighted in Section \ref{sec:Fair for GA}. Section \ref{sec:conclusion} present the conclusion and future guidelines. 	

\section{Preliminaries} \label{sec:Prelim}

\subsection{Genetic Algorithm (GA)}
GA is an evolutionary algorithm that has gained much importance in the last few decades due to its simplicity and effectiveness for complex optimization problems. GA is a directed randomization technique based on Charles Darwin’s theory of “Natural’s Selection” \cite{chambers2019practical,mitchell1998introduction}. Randomization helps GA to avoid local optima while the directed approach helps to converge to an optimal solution. GA uses stochastic operators (i.e. crossover and mutation) that helps to explore the search space and exploit the solutions respectively. GA starts by initializing a population of candidate solutions. Each candidate solution represents a string of feature/decision variables. 
The population is evolved by applying GA operators on the candidate solutions. The fitness of the candidate solution is evaluated using a fitness function that is mainly problem dependent. The termination criteria is based on the maximum number of generations, the maximum amount of time, or the specified convergence criteria. Different variants of GA (i.e Sequential GA, Parallel GA and Distributed GA) are briefly explained in Table \ref{tab:Variants of GA}. GA has different population initialization methods (i.e Random, Feasible Individuals, and Random and Greedy) as explained in Table \ref{tab:Initial}. Moreover GA has also different population structures and how individual solutions communicate with each other as shown in Table \ref{tab:Population}. 

\begin{table*}[!htb]
\caption{Different variants of GA\label{tab:Variants of GA}}
\renewcommand{\arraystretch}{1.1}
\centering
\scriptsize
\begin{tabular}{|p{0.05\linewidth}|p{0.15\linewidth}|p{0.5\linewidth}|p{0.15\linewidth}|}
\hline
S.NO & GA Variant & GA Variant Detail & Reference \\
\hline
1 & {Sequential GA} & It starts with a single population of solutions and evolves it over time by applying GA operators. The process continues until the desired convergence, or required generations are reached. & \cite{chambers2019practical, mitchell1998introduction}  \\
\hline
2 & {Parallel GA} & The initial population is divided into subpopulations. Multiple GA operations are performed in parallel like fitness evaluation, selection, crossover, and mutation. & \cite{liu2015scalable, trivedi2015hybridizing,lim2007efficient, ferrucci2018using, qi2016parallel,cantu1998survey}  \\
\hline
3 & {Distributed GA} & In this variant, dimensions of individuals or population are distributed. For dimension distribution multi-agent and coevolution methods are used. In population distribution Island, Hierarchical, Master slave, Cellular, and Pool model are used. & \cite{maqbool2019scalable, roy2009distributed,eklund2004massively,folino2008training, akopov2013multi, cao2017distributed, dubreuil2006analysis, sefrioui2000hierarchical,  alba2005advanced, artyushenko2009analysis} \\

\hline
\end{tabular}
\end{table*}

\begin{figure}[!htb]
\centering
\includegraphics[width=0.98\textwidth]{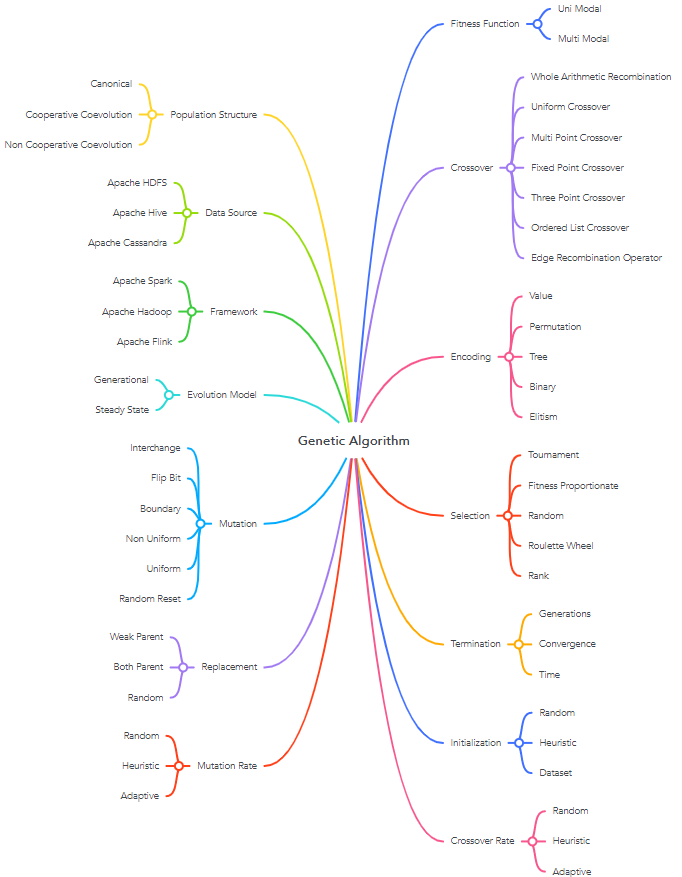}
\caption{Detailed metadata parameters  to improve the reusability and reproducibility of GA. }
\label{fig:detailed-metadata}
\end{figure}

While working with GA, researchers must carefully select and specify essential parameters like population initialization, population structure, encoding scheme, selection criteria, crossover technique, crossover rate, mutation rate, mutation technique, and replacement criteria as shown in detail in Figure \ref{fig:detailed-metadata}. The suggested details helps the researchers to express GA metadata more appropriately and use the existing GA techniques to reproduce the results effectively.

\begin{table*}[!tb]
\caption{Population initialization methods in GA\label{tab:Initial}}
\renewcommand{\arraystretch}{1.1}
\centering
\scriptsize
\begin{tabular}{|p{0.05\linewidth}|p{0.15\linewidth}|p{0.5\linewidth}|p{0.15\linewidth}|}
\hline
S.NO & Population Initialization & Initialization details & Reference \\
\hline
{1} & Random & The initial population is randomly selected without any heuristic and constraint. & \cite{skok2000genetic,li2009research,salhi2007ga} \\
 
\hline
{2} &  {Feasible Individuals} & The initial population contains selected/possible individuals as an initial population set. & \cite{ghoseiri2010multi,alvarenga2007genetic} \\

\hline
{3} &  {Random and Greedy} & The initial population is based on a random and greedy mixed approach.  &  \cite{ombuki2006multi,santos2006combining}  \\

\hline
\end{tabular}
\end{table*}

\begin{table*}[!bh]
\caption{Population structures in different variants of GA\label{tab:Population}}
\renewcommand{\arraystretch}{1.1}
\centering
\scriptsize
\begin{tabular}{|p{0.05\linewidth}|p{0.12\linewidth}|p{0.58\linewidth}|p{0.09\linewidth}|}
\hline
S.NO & Population Structure & Description & Reference \\
\hline
1 & Conventional GA & A combined population pool where each individual can interact with any other individual. & \cite{lim2014structured} \\
\hline
2 & Island Model & The initial population is divided into multiple subpopulations/islands. On each island, GA operators work independently. & \cite{artyushenko2009analysis} \\
\hline
3 & Cellular Model & Each individual can interact within a defined small neighborhood, and GA operations are applied to them. & \cite{alba2005advanced} \\
\hline
4 & Terrain-Based & Parameters are available across the population. At each generation, an individual can interact with the best individual in a close neighborhood. & \cite{krink1999patchwork} \\
\hline
5 & Spatially-Dispersed & Once the first individual/parent is selected, the second individual is chosen based on its spatial coordinates visibility from the first individual/parent. & \cite{dick2003spatially} \\
\hline
6 & Multilevel Cooperative & The population is divided into multiple groups. Offsprings in sub populations evolve and are updated by replacing weak individuals. & \cite{reza2010mlga}\\
\hline

\end{tabular}
\end{table*}


We have also performed a limited survey to support our claim that most of the GA algorithms are not FAIR. We only selected top 50 articles against the keyword search\footnote{\url{https://scholar.google.com/scholar?as_ylo=2021&q=Genetic+Algorithm&hl=en&as_sdt=0,5} Accessed: 30-08-2022, 12:00} "Genetic Algorithm" from Google Scholar, published in year 2021. From 50 articles, only 03 articles \cite{weissbrich2021using, oliveira2021c++, hooft2021discovering}  mention the source code \footnote{\url{https://github.com/Ensing-Laboratory/FABULOUS}}, \footnote{\url{https://github.com/tubs-eis/VANAGA}}, \footnote{\url{http://mauricio.resende.info/src/coEvolBrkgaAPI}} of their proposed technique. Moreover most of the GA-based code repositories available on Github \footref{1st},\footref{2nd},\footref{3rd},\footref{4th},\footref{5th},\footref{6th},\footref{7th},\footref{8th},\footref{9th},\footref{10th},\footref{11},\footref{12},\footref{13},\footref{14} do not provide the required hyper-parameter settings, configuration parameters, and metadata details. We reiterate here that the purpose of this survey is neither to perform a exhaustive overview of Genetic Algorithm, nor to provide a thorough understanding of the algorithm. Rather it is aimed  to merely highlight the main challenges in findability and reproducibility of GA research software by selecting a sample. Another challenge to the findability is multiple naming conventions of GA and its variants available in literature \cite{oliveira2021c++, hooft2021discovering}. Researchers may intermix multiple conventions leading to poor findability.

\subsection{FAIR}
The journey of making data FAIR (Findable, Accessible, Interoperable, Reusable) data started from the guidelines initially proposed by Wilkinson et al. \cite{wilkinson2016FAIR} enlisted in Table \ref{tab:FAIR for data}. Research communities were suggested to follow the guidelines and agree upon common data and metadata storage framework. FAIR not only helps the researchers to get the maximum potential (by attracting new research partnerships and increasing  citations/visibility) from the data set. It also helps in to improve the reusability and reproducibility of the data by building novel resources and tools by taking maximum benefits from the existing datasets. FAIRification of data is based on four key components i.e Findable, Accessible, Interoperable, and Reusable.\\

Findability suggests that the necessary practices that should be carried out to make your data and metadata easily findable by both man and machines. In order to ensure data availability, it should be placed in such a way that every element of data and metadata is accessible using a unique and persistent URI. This will help to avoid the ambiguity about the elements. Search engines are the key source of information nowadays. In order to make data/metadata findable by the search engines, it should not only be placed on index-able sources so that search engine bots may read and index them in their SERP (Search Engine Result Pages) but also metadata should be rich enough so that it may contain the necessary information/keywords based on which you want to be found in search engines. Including data identifiers in metadata file will help increase clarification about the data for which metadata is being defined.

Accessibility doesn't state that data should be freely accessible to everyone but rather there should be a clear machine readable guidelines available that states who can access the data. Data accessibility using standard communications protocols (e.g. HTTP, FTP, SMTP) not only ensures data reusability but also help to increase this if the protocols are available free of cost. Data storage carries a cost and may become unavailable overtime. Metadata should even remain available in case of broken links to dataset so that if someone interested in the dataset then he/she may track the publisher or author of the dataset using the metadata.\\
To ensure the interoperability, a formal, shared, and broadly applicable, metadata format is required, that follows standard vocabularies and qualified references to other metadata. Interoperability helps machine to understand exchanged data format from other machines with the help of standardized ontology and vocabularies. 
In reusability, metadata is enriched with detailed attributes about the data along with a clear data usage license. Complete details and conditions under which data was generated through experiments/sensors/machine/protocol and citation details is mentioned in metadata. In this article we have adapted and extended FAIR data principles for algorithms(FAIR-Algorithms). Later we have presented a usecase by applying FAIR-Algorithms using GA.

\section{Related Work} \label{sec:Related Work}
Researchers usually follow different steps in their research process ranging from problem analysis, literature review, data collection, research software development/usage, experiments, and result analysis. By following FAIR guidelines, researchers can easily reuse published data, research software, and results. It also helps to increase the focus on extending the existing work and achieving their research goals at rapid pace. Recently, efforts have been made by research community to support the FAIR research culture. Recent development in academia \cite{wilkinson2016FAIR}, Life Science \cite{vogt2019FAIR}, FAIR ontologies \cite{guizzardi2020ontology},  SmartAPI \cite{vita2018FAIR}, Immune Epitope Database \cite{spoor2019tripal} and Health Care \cite{sinaci2020raw}, have been made to make data, webAPI, ontologies and biological databases FAIR.\\

Academia and publishing industry must play an active and vibrant role in making such efforts to publish data, research software, and research manuscripts according to FAIR principles. In this regard few journals (JORS, IPOL, JOSS, eLife, and science direct) have already started to review the research software during the peer review process and publish it along with the article \cite{gruenpeter2020m2}. Also Association for Computing Machinery (ACM), has started to review research software, datasets, experiments, and other related files, along with research manuscripts \cite{krishnamurthi2015real}. ACM has introduced the policy badges to review the research articles for results replication i.e. (same results of an article) or results reproduction (results generated independently) mechanism. ACM also encouraged reproducibility, reusability, and replicability in which the experimental setup of software can be used or extended by the same or a different team \cite{boisvert2016incentivizing}. Moreover few recommendations for executing FAIR practices including (training, education, fundraising, incentives, rewards, recognition, development, and monitoring of policies) were suggested by Hong et.al. \cite{hong2020six}. Also 
FAIR has far reaching benefits for different domains. In  agriculture, Basharat et.al \cite{ali2022role} discussed role of FAIR data in agriculture industry. They have applied FAIR guidelines to ensure data findabilty and reusabilty of agriculture data, used for decision making and agriculture performance. For making a molecular plant data FAIR a check list is compiled by Reiser et.al \cite{reiser2018fair}, it includes placing data at a stable repository, using unique identifiers for genes and its products, by using standard file formats, reproducible computational technique by mentioning ( software versions, raw data files, citing data source, parameter settings). Different industries have started their projects of making data FAIR \cite{van2020need}. A project on making life science data FAIR  FAIRplus \footnote{https://fairplus-project.eu/} is in progress.

FAIR guidelines are independent of tools, technologies, and implementation platforms \cite{wilkinson2016FAIR}. There are few common and some exclusive details for FAIR data and FAIR research software’s suggested by Lamprecht et al. \cite{lamprecht2020towards}. They have adopted some of the existing FAIR principles where they fit in for research software’s and modify/extend the remaining one. The list of recommendations for FAIR research software based on existing FAIR guidelines for data is proposed by Hasselbring et al. \cite{hasselbring2020FAIR}. Software development community also suggested that FAIR research software principles should be separately defined \cite{gruenpeter2020m2}. There are different challenges ( i.e. software documentation, accessibility, licensing issues, software dependencies, environment, quality control, and software sustainability) in making research software findable and reusable \cite{org2020d2}.
All these efforts are made in recent years for making scientific data, software, and related objects FAIR. Different digital artifact repositories (i.e. Github\footnote{\url{https://guides.github.com/features/pages/},Accessed Oct 8,2021}, GitLab\footnote{\url{https://gitlab.com/gitlab-org/gitlab}, Accessed Oct 8,2021}, Zenodo\footnote{\url{https://zenodo.org/},Accessed Oct 8,2021},SourceForge\footnote{\url{https://sourceforge.net/},Accessed Oct 8,2021},and Bitbucket\footnote{\url{https://bitbucket.org/},Accessed Oct 8,2021}) are used to store and publish the data and software. 
The findability of data and research software is enforced by the relevant conference, workshop, or journal at the time of publishing of manuscript. Similarly, the accessibility of data and research software has its own challenges. Data and software usage license and copyrights details should be clearly stated and permission of access should be granted to the research community where admissible. Reproducability of research software is also a challenge and this is due to the lack of availability of software code, its Persistent IDs, and reproducing the complete software environment as highlighted by Alliez et al.\cite{alliez2019attributing}. To improve the reusability by the research community for the photovoltaic time series data, a set of recommendation's  were suggested by Arafath et.al. \cite{nihar2021toward}. It includes clearly defined dataset, accessibility and availability of metadata in human and machine readable format i.e JSON-LD. 

Another challenge related to research software is not a well-defined attribution mechanism for the developers of the research community. This results in less focus on quality research software but on research manuscripts~\cite{lamprecht2020towards}. Current citation mechanisms, impact factor policies, and promotion/hiring in universities are research publication centric. Preliminary work on software citation principles was highlighted by Smith et al.\cite{smith2016software}. They have identified, importance, credit, attribution, unique identification, persistence, accessibility, and specificity as the major software citation principles. Format of citing software, metadata of software for citation, criteria for peer review of the software, and acceptance of software as a digital product were the few challenges related to software citations that were highlighted by Niemeyer et al. \cite{niemeyer2016challenge}. The research community, journals, conferences, workshops, and research and project funding agencies have to initiate such reforms that help in making research data, software, and related research objects FAIR. All the relevant stakeholders have to develop/encourage practices, like citation incentives and scholarly attribution for research software developers/data analysts/researchers for following FAIR principles.\\
To the best of our knowledge, we are unable to find any application of FAIR guidelines to algorithms, so in this article, we have extended FAIR data guidelines to develop FAIR-Algorithms. Moreover to justify the applicability of FAIR-Algorithms we have presented a use case (i.e FAIR-GA) and validated our $FAIR-Algorithms$ proposed guidelines.  

\section{FAIR-Algorithms: FAIR Principles for Algorithms} \label{sec:FAIR Principles}
An algorithm is a set of instructions to complete a specific task.
It is usually designed to solve a specialized problem
/ sub-problem and consists of inputs, tasks, outputs, and parameters settings. Inputs are the data and parameters, while the task is the main description of the work that uses inputs to generate outputs(reports, computational outcome, models).\\ 
Sharing the data, research software, algorithm and related metadata is required in improving, reusing, or reproducing algorithms with a purpose to improve efficiency, efficacy, or resource utilization. Hence a recent trend of developing FAIR principles for software may be of interest to those researchers who view software as a black box or an atomic entity. This motivated us to work on FAIR guidelines for algorithms. \\
The idea behind FAIR-Algorithms is that if a research is FAIR then not only others should be able to reproduce data and software but also should be able to reproduce, reuse, extend, or build on top of the algorithm. \\
We have enlisted FAIR principles for algorithms in Table \ref{tab:FAIR for Algo} and mentioned the action (i.e. adapted, and extended) against each principle. Adapted is used where FAIR data principle is used for algorithm with out any modification and extended is used when existing FAIR principles is modified to cover the algorithm and its related details. FAIR-Algorithms guidelines are presented in Table \ref{tab:FAIR for Algo}.\\



\begin{table*}[!t]
\caption{FAIR principles for Algorithms\label{tab:FAIR for Algo}}
\renewcommand{\arraystretch}{1.1}
\centering \scriptsize
\begin{tabular}{|p{0.05\linewidth}|p{0.05\linewidth}|p{0.67\linewidth}|p{0.1\linewidth}|}
\hline
FAIR & ID & FAIR for Algorithms & Action\\
\hline
\multirow{4}{*}{F} & 1 & Algorithm and its metadata is assigned a globally unique and persistent identifier. & extended \\
\cline{2-4}
 & 2 & Algorithms are described with rich metadata. & adapted \\
\cline{2-4} 
 & 3 & Algorithm metadata clearly and explicitly include the identifier of the algorithm it describes. 
    & extended \\
\cline{2-4}
 & 4 & Algorithm metadata are indexed on a searchable repository. & extended \\
\hline

\multirow{4}{*}{A} & 1 &  Algorithm metadata are retrievable by their identifier using a standardized communications protocol. & adapted\\
\cline{2-4}
 & 1.1 &  The protocol is open, free, and universally implementable. & adapted\\
\cline{2-4}
 & 1.2 &  The protocol allows free and easy access to algorithms' metadata and details.& extended \\
\cline{2-4}
 & 2 & Algorithm-metadata remains available, even when the algorithms are modified. & extended \\
\hline

\multirow{3}{*}{I} & 1 & Algorithm metadata uses a formal, accessible, shared, and broadly applicable language for knowledge representation.& adapted  \\
\cline{2-4}
 & 2 & Metadata uses vocabulary that follows FAIR principles. & adapted  \\
 \cline{2-4}
 & 3 & Algorithm metadata include qualified references to other metadata. & adapted \\
\hline

\multirow{4}{*}{R} & 1 & Algorithm metadata is richly described with a plurality of accurate and relevant attributes. & adapted  \\
\cline{2-4}
 & 1.1 & Usually, algorithm metadata is freely accessible. However, if required, algorithm metadata is released with a clear and accessible usage license. & extended\\
\cline{2-4}
 & 1.2 & Algorithm metadata includes detailed provenance. It includes its basics, execution and performance attributes & extended  \\
\cline{2-4}
 & 1.3 & Algorithm metadata meet domain-relevant community standards. & adapted \\
\hline

\end{tabular}
\end{table*}

\subsection{Findability}
\textbf{F1:- Algorithm and its metadata is assigned a globally unique and persistent identifier.}\\
Existing digital artifact repositories (i.e. Github\footnote{\url{https://github.com/}},
GitLab\footnote{\url{https://gitlab.com/}}, Zenodo\footnote{\url{https://zenodo.org/}},  
SourceForge\footnote{\url{https://sourceforge.net/}}, 
Bitbucket\footnote{\url{https://bitbucket.org/}}) do not assign a unique identifier to the algorithm, rather to a software repository that may contain one or more algorithms. Therefore it is recommended that there should be a unique metadata file related to each algorithm. Also a unique identifier should be assigned to each metadata file and algorithm. An algorithm has a unique identifier, 
If its sub algorithms have no unique ID (UID), and the owner wants to assign UID, he can opt to do so. On the other hand, if an algorithms is used in another algorithms it UID will be used, and the new/parent algorithms will be assigned a new UID. We do not imagine allocation of UIDs retrospectively. Moreover different implementation of same algorithms have different UIds and in case of updates in an implementation its versioning control should also be maintained.\\

\textbf{F2:- Algorithms are described with rich metadata}\\
The metadata for an algorithm includes its input, tasks/steps, output, implementation details, parameter settings, execution environment, and execution duration. We have extended the MEX vocabulary \cite{esteves2015mex} to define the metadata for algorithms as given in Table \ref{tab:spec-meta for Algo}. Algorithm's metadata file also includes the UID of the algorithm in it.

\textbf{F3:- Algorithm metadata clearly and explicitly include the identifier of the algorithm it describes.}\\

Currently there is no practice of publishing algorithm's metadata. However we have recommended that algorithm's metadata should clearly and explicitly point to the algorithm that is being described, these identifiers include the identifier, author, usage information, citation, and other related properties as suggested in metadata for algorithms in Table \ref{tab:spec-meta for Algo}. 

\textbf{F4:- Algorithm metadata are indexed on a searchable repository}\\
Publishing algorithm's metadata is not in practice and hence not indexed. Therefore, we have suggested that algorithm's metadata should be placed on digital artifact repositories (i.e Zenodo, Github, GitLab, and BitBucket) that quickly index the published resources and make them searchable.\\

\subsection{Accessability}
\textbf{A1:- Algorithm metadata are retrievable by their identifier using a standardized Communications protocol.}\\
Existing artifact repositories (i.e Zenodo, Github, GitLab, and BitBucket) are accessible using standard communication protocols like http/https. So, metadata placed on these repositories is also accessible. Hence, we recommend to use these for algorithms.  \\

\textbf{A1.1:- The protocol is open, free, and universally implementable.}\\
Generally, the https protocols are open, free, and universally used. The algorithms that are published on above mentioned digital artifact repositories are using these free access protocols. In case of private publishing of Algorithms or its metadata, the metadata must be made accessible through universally acceptable protocols.\\

\textbf{A1.2:- The protocol allows free and easy access to algorithms’ metadata and details.}\\

Mostly algorithms don't have authentication and authorization issues. In case of privacy related issues in publishing a particular algorithm, the metadata should still remain accessible, even after following an authentication and authorization procedure. However, the authentication and authorization procedure may be adapted where necessary.


\textbf{A2:- Algorithm-metadata remains available, even when the algorithms are modified.}\\
New variants of algorithms are proposed over time and older variants may reduce their public visibility. Therefore, metadata for the earlier versions of an algorithm should remain accessible and available even after its new version or extension has become more prevalent.\\

\subsection{Interoperability}
\textbf{I1:- Algorithm metadata uses a formal, accessible, shared, and broadly applicable language for knowledge representation}\\
Current artifacts repositories support XML\footnote{\url{https://www.w3.org/XML/}}, JSON\footnote{\url{https://www.json.org/}}, JSON-LD\footnote{\url{https://json-ld.org/}}, and Rest APIs\footnote{\url{https://restfulapi.net/}} as broadly applicable languages. Therefore we recommend to use above mentioned broadly applicable formats for algorithm's metadata. 

\textbf{I2:- Metadata uses vocabularies that follow FAIR principles.}\\
Fairification of vocabularies to define algorithm's metadata have been under explored. In this regards, we have suggested to use and extend (where necessary) MEX Vocabulary (comprising $mexcore$\footnote{\label{core}\url{https://github.com/mexplatform/mex-vocabulary/blob/master/vocabulary/mexcore.ttl}}, $mexalgo$\footnote{\label{algo}\url{https://raw.githubusercontent.com/mexplatform/mex-vocabulary/master/vocabulary/mexalgo.ttl}}, $mexperf$\footnote{\label{perf}\url{https://github.com/mexplatform/mex-vocabulary/blob/master/vocabulary/mexperf.ttl}}) \cite{esteves2015mex} for algorithm metadata as it maximally satisfies FAIR principles. $mexcore:Context$, $mexalgo:AlgorithmClass$, $mexcore:model$, $mexalgo:AlgorithmParameter$, and $mexalgo:Implementation$ are few of the main classes of the vocabulary.

\textbf{I3:- Algorithm metadata include qualified references to other metadata.}\\
Currently there is no practice of publishing algorithms metadata. Once it is started, focus on referencing other related metadata would also be in practice. It will help in increasing algorithm reusability.


\subsection{Reusability}
\textbf{R1:- Algorithm metadata is richly described with a plurality of accurate and relevant attributes.}\\
Rich metadata is helpful in understanding an algorithm. We have combined metadata specifications in detail in Table \ref{tab:spec-meta for Algo}. It includes basic attributes from schema\footnote{\url{https://schema.org/}}. Algorithm, parameters, learning methods, tool, class from $mexalgo$. Performance measure and user defined measures from $mexperf$. All these details $mexcore$, $mexalgo$ and $mexperf$ are taken from mex vocabulary \cite{esteves2015mex}. Algorithm metadata is richly described by following these detailed set of attributes.\\
\textbf{R1.1:- Usually, algorithm metadata is freely accessible. However, if required, algorithm metadata is released with a clear and accessible usage license.}\\
Copyright protects the creative work or expression of ideas but not the ideas themselves. An algorithm is an abstract idea and not subject to licensing \cite{cormen2009introduction}. However, the code of the algorithm should have licensing information. An algorithm code is part of the research software, so the algorithm's usage license is as per the discretion of the research software team.\\

\textbf{R1.2:- Algorithm metadata includes detailed provenance. It includes its basics, execution and performance attributes}\\
Currently metadata specification for algorithm is not in practice. We have specified metadata specifications in detail in Table \ref{tab:spec-meta for Algo}. These metadata specifications helps in making algorithm FAIR.

\textbf{R1.3:- Algorithm metadata meet domain-relevant community standards.}\\
To the best of our knowledge, there doesn't exist any algorithm relevant community standards. Therefore it is suggested to follow the guidelines from algorithm related vocabularies and ontologies.

\begin{figure}[thpb]
\centering
\includegraphics[width=1\textwidth]{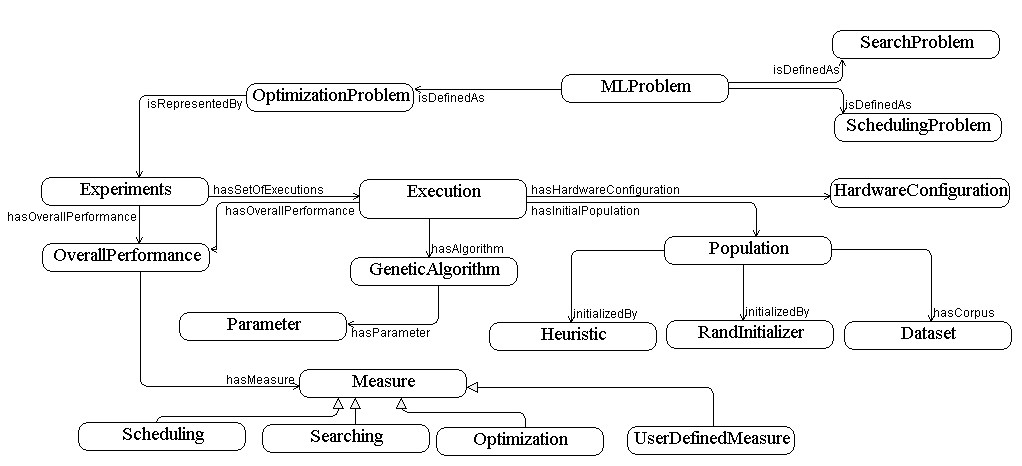}
\caption{GA iteration cycle starting from the problem specification till the performance measures.}
\label{fig:GAParameters}
\end{figure}

\begin{figure}[thpb]
\centering
\includegraphics[width=1\textwidth]{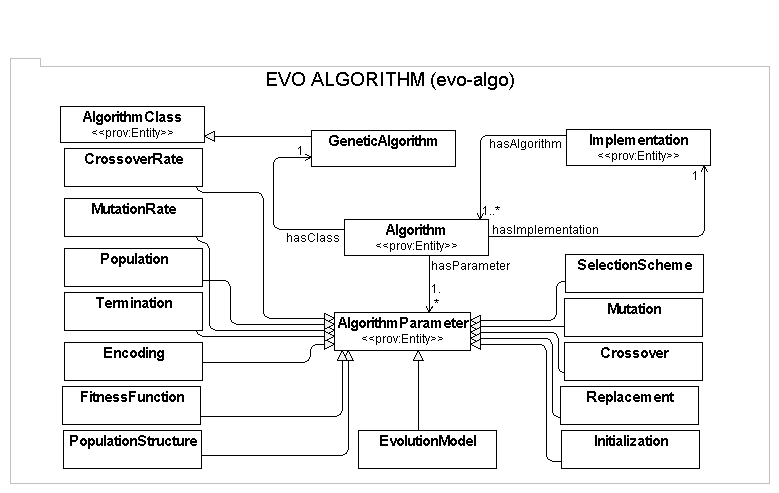}
\caption{Algorithm related parameter specification for GA}
\label{fig:detailed-metadatainfo}
\end{figure}

\begin{figure}[bth]
\centering
\includegraphics[width=1\textwidth]{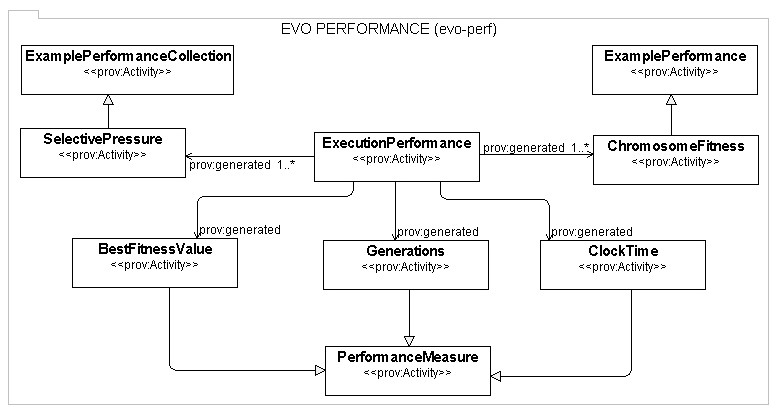}
\caption{Performance related parameter specification for GA}
\label{fig:detailed-metadataperf}
\end{figure}

\section{Metadata for Genetic Algorithm} \label{sec:GA-Metadata}

Rich metadata of a digital artifact specified using a well known data format plays an important role in machine readability. Also it plays a vital role in reusability and reproducibility using well defined hyper-parameter values. Algorithm metadata based on mex-vocabulary covers all general-purpose attributes(like tools, dataset, feature etc) that are common among algorithms. Defining GA specific attributes (like population size, fitness function, crossover rate) is not workable using mex-vocabulary and demands an extension of attributes in mex-vocabulary.
In this section we have have suggested specific metadata \footref{mt} for GA as listed in table \ref{tab:spec-meta for GA}. Listing \ref{listing:ga} shows experiment metadata of a Python based GA for solving one max search optimization problem \footnote{\url{https://colab.research.google.com/drive/1t_kUu6l3a4F1sP6oK92CHZwqANsTMijP?usp=sharing}}. We have represented this metadata using minimal classes and properties from prov-o\footnote{\url{https://www.w3.org/TR/prov-o/}}, mex-core\footnote{\url{http://mex.aksw.org/mex-core#}}, mex-algo\footnote{\url{http://mex.aksw.org/mex-algo#}}, and mex-perf\footnote{\url{http://mex.aksw.org/mex-perf#}}. 
Moreover we also suggest to improve upon MEX vocabulary and propose $evo$ vocabulary. A hierarchical representation of these parameters has been shown in Figure \ref{fig:detailed-metadata}. Few main terms/entities of $evo$ vocabulary are listed below.  
\begin{enumerate}[noitemsep]
  \item $evo:SelectionScheme$
  \item $evo:Encoding$
  \item $evo:FitnessFunction$
  \item $evo:Error/Loss Function$
  \item $evo:PopulationSize$
  \item $evo:Crossover$
  \item $evo:CrossoverPoint$
  \item $evo:CrossoverProbability$
  \item $evo:Mutation$
  \item $evo:MutationProbability$
  \item $evo:Replacement$
  \item $evo:Elitism$
  \item $evo:Termination$
  \item $evo:Generations$
  \item $evo:Time$
  \item $evo:Fitness$
\end{enumerate}

Figure \ref{fig:GAParameters} shows the problem specific parameters for GA. The detailed parameters includes the problem (i.e. scheduling, searching, or optimization), GA, population (initialized using dataset, random initializer, or heuristic). Algorithm related parameters for GA and performance related parameters for GA are shown in Figure \ref{fig:detailed-metadatainfo} and \ref{fig:detailed-metadataperf} respectively. 

{\scriptsize
\onecolumn
\renewcommand{\arraystretch}{1.1}
\centering
\begin{longtable}[!h]{ | p{0.31\linewidth} | p{0.3\linewidth}| p{0.3\linewidth} |}

\caption{Metadata specification for Algorithm.\label{tab:spec-meta for Algo}}    \\

\hline
\endfirsthead

\multicolumn{3}{|c|}{Metadata specification for Algorithm (Continued). }   \\
\hline
\endhead
\hline
\endfoot
\hline
\endlastfoot

\hline
Property & Description & Type  \\
\hline
schema:identifier & DOI of algorithm & {\scriptsize schema:URL }\\
 \hline
schema:url & URL of software repository & {\scriptsize schema:URL }\\
 
\hline
schema:name &  Name of the algorithm &  {\scriptsize schema:Text }\\
\hline
schema:description &  Algorithm description that discusses Algorithm purpose and application areas.  &  {\scriptsize schema:Text }\\
\hline

schema:author &  Person / organization that has created the code and holds its intellectual copyrights. & {\scriptsize schema:Organization \newline schema:Person }\\
\hline

schema:usageInfo &  Limitation of the algorithm. &  {\scriptsize schema:Text }\\
\hline

schema:keywords &  Keywords and tags that describe the key terms to define a software. &  {\scriptsize schema:Text }\\
\hline

schema:citation &  Source code attribution (i.e. link to the article where the particular algorithm has been discussed). & {\scriptsize schema:Text }\\
\hline

schema:license & It helps to protect the intellectual property by defining the guidelines for use and distribution of software. & {\scriptsize schema:URL }\\
\hline
mexcore:ApplicationContext &  Basic information about algorithm that may provide a high-level overview (including goals, aims, objectives, and scope) of the software &{\scriptsize mexcore:ApplicationContext \{ \newline :trustyURI rdfs:Literal \newline mexcore:trustyURIHash  \}  }\\
\hline

mexcore:Context & The problem for which algorithm has been employed e.g data clustering, neural network optimization, and protein folding. &{\scriptsize mexcore:Context \{\newline prov:wasAttributedTo mexcore:ApplicationContext \newline \} }\\
\hline

mexcore:Experiment & The class represents some basic information about the experiment. & {\scriptsize mexcore:Experiment \{ \newline mexcore: attributeSelectionDescription xsd:string \newline :dataNormalizedDescription xsd:string \newline :noiseRemovedDescription xsd:string \newline :outliersRemovedDescription xsd:string \newline  \} }\\
\hline

mexcore:Execution & A single run of an algorithm-based program. Each run is based on specific parameter specification and hardware configurations. & {\scriptsize mexcore:Execution \{ \newline mexcore:endsAtPosition xsd:string \newline :targetClass xsd:string \newline prov:wasInformedBy mexcore:ExperimentConfiguration \newline \}    }\\
\hline

mexcore:ExperimentConfiguration & It represents execution detail (on different algorithm configuration and hardware environments) of an experiment. & {\scriptsize mexcore: ExperimentConfiguration \{ \newline prov:wasStartedBy mexcore:Experiment \newline \}   }\\
\hline

mexcore:HardwareConfiguration & Detail about hardware configuration & {\scriptsize mexcore: HardwareConfiguration \{ \newline mexcore:cpu xsd:string \newline mexcore:cpuCache xsd:String \newline mexcore:hdType xsd:string \newline mexcore:memory xsd:string \newline :videoGraphs xsd:string \newline \} }\\
\hline


mexcore:DataSet & Initial population/ dataset for algorithm experiments & {\scriptsize owl:Class} \\
\hline

mexcore:Example & An individual solution or a chromosome & {\scriptsize   mexcore:Example \{ \newline mexcore:datasetColumn rdfs:Literal \newline mexcore:datasetRow rdfs:Literal \newline   \}      }\\
\hline

mexcore:ExampleCollection & ExampleCollection is a collection of chromosomes and represents a population at a particular generation. & {\scriptsize  mexcore:ExampleCollection  \{  \newline mexcore:startsAt rdfs:Literal \newline mexcore:endsAt rdfs:Literal \newline mexcore:hasPhase mexcore:Phase \newline   \}      }\\
\hline


mexalgo:LearningMethod & 
This defines the learning approach of the algorithm i.e. evolution in case of genetic algorithm. & {\scriptsize  mexalgo:LearningMethod \{ \newline mexalgo:isLearningMethodOf mexalgo:Algorithm \newline \}      }\\
\hline

mexalgo:LearningProblem & GA is a metaheuristic based algorithm & {\scriptsize  mexalgo:LearningProblem  \{ \newline :isLearningProblemOf :Algorithm \newline    \}      }\\
\hline

mexalgo:AlgorithmClass & The algorithm class (e.g.:GeneticAlgorithm) & {\scriptsize  mexalgo:AlgorithmClass  \{ \newline :isAlgorithmClassOf :Algorithm \newline    \}      }\\
\hline

mexalgo:AlgorithmParameter & The representation of GA parameter with its associated values (e.g. encoding, population, crossover scheme) & {\scriptsize  mexalgo:AlgorithmParameter  \{     \}      }\\
\hline  


mexalgo:Tool & It describes the libraries for GA (e.g.: PyGAD, GAlib, GeneAI). & {\scriptsize  mexalgo:Implementation  \{     \}      }\\
\hline

mexperf:PerformanceMeasure & It describes the evaluation measure to check the performance of GA (e.g.: Fitness function). &  {\scriptsize  mexperf:PerformanceMeasure  \{ \newline      \}      }\\
\hline

mexperf:UserDefinedMeasure  & This property is used to mention domain relevant metrics.       &   {\scriptsize   mexperf:UserDefinedMeasure \{\newline mexperf:formula xsd:string \newline    \}      }\\
\hline

\end{longtable}
}

\begin{table}[H]
{\scriptsize
\caption{Proposed $evo$ vocabulary for GA
\label{tab:spec-meta for GA}}
\renewcommand{\arraystretch}{1.1}
\centering
\begin{tabular}{ | l | p{0.6\linewidth}| p{0.1\linewidth} |}
\hline
Property & Description & Type  \\
\hline

evo:Initialization & Their are different initilization methods (i.e random, heuristic and dataset). & schema: Text \\
\hline

evo:Encoding & GA works on the encoding of the solutions rather than solutions. These may include binary encoding, value encoding, and permutation encoding.  & schema: Text  \\
\hline

evo:Bound & It represents the upper and lower bound values for each dimension in the state space.  & schema: Text\\
\hline

evo:PopulationSize & Population size is generally dependent on the problem domain and may be decided by hit and trial method. It has a significant role to speed up the convergence. & schema: Number \\
\hline

evo:FitnessMeasure & It is used to represent the final fitness value. & schema: Number \\
\hline

evo:TimeMeasure & This represents the clock time used by the GA & schema: Duration \\
\hline

evo:GenerationMeasure & It represents the total generation consumed during the execution. & schema: Number \\
\hline

evo:Evolution & It represents the learning method used by the GA & schema: Text\\
\hline

evo:Crossover & Parent chromosomes recombine to create new offsprings. Crossover property is used to specify the crossover operator (i.e. single point crossover, multi-point crossover, uniform crossover, or three parent crossover). & schema: Text  \\
\hline

evo:CrossoverRate & It is used to to decide the number of parents involved in the crossover process & schema: Text \\
\hline
evo:Mutation & Mutation operator helps to avoid getting stuck in local optima by maintaining the population diversity. Popular mutation includes bit flip, inversion, scramble, and random resetting. & schema: Text  \\
\hline

evo:MutationRate & It is the frequency measure by which value of randomly selected genes would be modified & schema: Text \\
\hline

evo:Selection & Selection scheme specifies the criteria through which parent chromosome will be selected from the current generation to produce offsprings using crossover and mutation. This may include rank selection, roulette wheel selection, and tournament selection. & schema: Text  \\
\hline

evo:PopulationUpdate & It can be steady state (i.e. off springs would be added to the population as they are created) or generational (i.e. off springs would be added to the population after the generation)  & schema: Text\\
\hline

evo:Replacement & Replacement is the scheme/strategy (weak parent, both parent, random parent) through which parent chromosomes will be replaced by newly created offsprings. & schema: Text  \\
\hline

evo:Termination & Termination specifies the criteria (i.e. generations, time, fitness, convergence) that stops the execution of genetic algorithm. & schema: Text  \\
\hline

evo:MaxGenerations & It specify the maximum number of iterations after which genetic algorithm will be terminated & schema: Number  \\
\hline

evo:FitnessFunc & It is used to mention the fitness function name (e.g. Sphere, Ackley or Griewank). Fitness function takes a solution as input and evaluates how close a given solution is to the optimal solution.   & schema: Text  \\
\hline

evo:FitnessFuncDef &  This class would be used to mention the fitness formula or fitness function definition. & schema: Text \\
\hline





evo:Time & Maximum amount of clock time after which we terminate the execution of genetic algorithm. & schema: Duration  \\
\hline

\end{tabular}
}
\end{table}

\section{$FAIR-GA$: FAIR Genetic Algorithm (A usecase of FAIR-Algorithms)}  \label{sec:Fair for GA}
Genetic algorithm is a popular optimization algorithm that is very effective in solving complex optimization problems. It initially starts with a population of encoded solutions, evolve these solutions using stochastic reproduction operators (i.e. crossover \& mutation), evaluate solutions using fitness function, eliminate less fit solutions, and proceeds with fittest solutions to next generations until termination criteria is met. Although GA and its variants have proved their significance in many optimization problems but lack of focus on reproducibility limits reusability of these techniques. In this section we have presented FAIR-GA (i.e. a use case of FAIR-Algorithms). This will not only help in supporting FAIR research culture but also helps researchers to increase their citation index and reusability of their proposed GA variants. Also, we have suggested guidelines that should be performed while mapping FAIR-Algorithms on GA (i.e. $FAIR-GA$).  These steps relate to the application of $FAIR-Algorithms$ on GA with some customization in F2, I2, R1, and R1.3 as discussed below.

\textbf{F2:- GA are described with rich metadata}\\
Taking care of FAIR standards, we suggest that GA metadata should contain detailed information about the input, output, algorithm name, algorithm task, parameters, parameters settings, citation detail, hardware details, and software dependencies. Algorithms metadata as listed in Table \ref{tab:spec-meta for Algo} are not inlined with the vocabulary requirements related to GA. GA has more specialized attributes (like population initialization, population size, solution encoding, generations, and termination criteria). Therefore proposed $evo$ vocabulary illustrated in Table \ref{tab:spec-meta for GA}  has to be collectively used with metadata specifications for algorithm described in Table \ref{tab:spec-meta for Algo} to improve the reusability and reproducibility of GA.

\textbf{I2:- GA software use vocabularies that follow FAIR principles}\\
For Fair-GA, we have proposed $evo$ vocabulary to support GA. The $evo$ vocabulary comprises of twenty properties as illustrated in table \ref{tab:spec-meta for GA} along with the description and type of each property. \\

\textbf{R1:- Metadata are richly described with a plurality of accurate and relevant attributes}\\
GA metadata must include accurate and relevant attributes. We have proposed extended metadata for GA as shown in Figure \ref{fig:detailed-metadata} and listed in $evo$ vocabulary given in table \ref{tab:spec-meta for GA}. This is a set of relevant and related attributes and suggested to be the part of GA techniques.\\

\textbf{R1.3:- Metadata meet domain-relevant community standards}\\
Our proposed $evo$ vocabulary is developed by looking into different state of the art GA variants. Detailed metadata has been suggested by taking care of all GA related hyperparameters configurations. However, we are unable to find any domain-relevant community standard for GA. 

In Listing \ref{listing:ga} we have applied, proposed $evo$ vocabulary along with other suggested FAIR-Algorithms guidelines for representing the experiments of GA one max search optimization problem.

\begin{lstlisting}[ language=json, firstnumber=1, caption=RDF representation of GA experiments for solving one max search optimization problem using MEX vocabulary and $evo$ vocabulary, label=listing:ga]
{
  {
  "@context": {
  "prov": "http://www.w3.org/ns/prov#",
  "mexperf": "http://mex.aksw.org/mex-perf#",
  "mexcore": "http://mex.aksw.org/mex-core#",
  "mexperf": "http://mex.aksw.org/mex-algo",
    "evo": "http://mex.aksw.org/evo"
  },
  "@id": "mexperf:ExecutionPerformance",
  "prov:generated": [	
    {
    "@id": "evo:FitnessMeasure",
    "evo:hasFitness": "-20"
    },
    {
      "@id": "evo:TimeMeasure",
      "evo:elapsedTime": "3",
      "evo:timeUnit": "nsec"
    },
    {
      "@id": "evo:GenerationMeasure",
      "evo:generationCount": "8"
    }    
                     ],


  "prov:wasInformedBy": {
    "@id": "mexcore:Execution",
    "prov:wasInformedBy": {
      "@id": "mexcore:ExperimentConfiguration",
      "prov:used": {
          "@id": "mexcore:HardwareConfiguration",
          "mexcore:hardDisk": "108GB",
          "mexcore:memory": "36GB"
        },
      "prov:wasStartedBy": {
      "@id": "mexcore:Experiment",
      "prov:wasAttributedTo": {
      "@id": "mexcore:ApplicationContext"
      }
      }
    },
    "prov:used": {
    "@id": "mexalgo:Algorithm",
      "schema:identifier": "https://doi.org/10.5281/zenodo.7096663",
      "schema:name": "Simple Genetic Algorithm",
      "schema:description": "A python implementation of a simple genetic algorithm to optimize the numerical functions.",
      "schema:author": [{
        "@id": "schema:Person",
        "name": "Saad Razzaq",
        "email": "saad.razzaq@uos.edu.pk"
      },
      {
        "@id": "schema:Person",
        "name": "Fahad Maqbool",
        "email": "fahad.maqbool@uos.edu.pk"
      },
      {
        "@id": "schema:Person",
        "name": "Hajira Jabeen",
        "email": "hajira.jabeen@gesis.org"
      }],
      "schema:keywords": "Evolutionary Optimization; Mutation; Crossover",
      "schema:license": "https://www.gnu.org/licenses/gpl-3.0-standalone.html",
      
      "mexalgo:hasClass": {
        "@id": "mexalgo:GeneticAlgorithms"
      },
      "mexalgo:hasLearningProblem": {
        "@id": "mexalgo:MetaHeuristic"
      },
      "mexalgo:hasLearningMethod": {
        "@id": "evo:Evolution"
      },
      "mexalgo:hasTool": {
        "@id": "mexalgo:Python"
      },
      "mexalgo:hasHyperParameter": [
        {
          "@id": "evo:Initialization",
          "prov:value": "Random"
        },
        {
          "@id": "evo:Bound",
          "prov:value": ["-5","+5"]
        },
        {
          "@id": "evo:Encoding",
          "prov:value": "Bit-String"
        },
        {
          "@id": "evo:PopulationSize",
          "prov:value": "100"
        },
        {
          "@id": "evo:Dimensions",
          "prov:value": "20"
        },
        {
          "@id": "evo:Crossover",
          "prov:value": "One-point Crossover"
        },
        {
          "@id": "evo:CrossoverRate",
          "prov:value": "0.9"
        },
        {
          "@id": "evo:Mutation",
          "prov:value": "Bit Flip"
        },
        {
          "@id": "evo:MutationRate",
          "prov:value": "1.0 / (float(n_bits) * len(bounds))"
        },
        {
          "@id": "evo:Selection",
          "prov:value": "Tournament"
        },
        {
          "@id": "evo:PopultionUpdate",
          "prov:value": "Generational"
        },
        {
          "@id": "evo:Replacement",
          "prov:value": "BothParent"
        },
        {
          "@id": "evo:Termination",
          "prov:value": "Generations"
        },
        {
          "@id": "evo:MaxGenerations",
          "prov:value": "100"
        },
        {
          "@id": "evo:FitnessFunc",
          "prov:value": "One-Max"
        },
        {
          "@id": "evo:FitnessFuncDef",
          "prov:value": "def onemax(x): return -sum(x)"
        }
      ]
    }
  }
}
}
\end{lstlisting}

\section{Conclusion} \label{sec:conclusion}

We have extended FAIR principles beyond data, so that these could be applied to methods, algorithms and software artifacts. Our focus in this article is to ensure the reproducibility and reusability of algorithms. We have presented $FAIR-GA$ as a usecase to demonstrate the applicability of the proposed principles for algorithms. Additionally, to ensure $FAIR-Algorithms$ we propose a metadata schema using light weight RDF format, facilitating the reproducibility. Finally, we have demonstrated the application of proposed $FAIR-GA$ using a python based GA code for solving one max search optimization problem. Moreover, we have also proposed a specialized vocabulary (i.e $evo$) for GA. In future this work can be extended to numerous machine learning algorithms by suggesting specialized vocabulary for making them FAIR.

\bibliographystyle{elsarticle-num-names} 
\bibliography{main}






\end{document}